\documentclass[runningheads]{llncs}
\usepackage[width=122mm,left=12mm,paperwidth=146mm,height=193mm,top=12mm,paperheight=217mm]{geometry}
\usepackage{graphicx}
\usepackage{amsmath}
\usepackage{amssymb}
\usepackage{color}
\usepackage{todonotes}
\usepackage{cleveref}
\usepackage{subcaption} 
\usepackage{booktabs}
\usepackage{tablefootnote}
\usepackage{subcaption}

\newcommand{\ra}[1]{\renewcommand{\arraystretch}{#1}}

\newcommand{\bp}{\mathbf{p}}
\newcommand{\br}{\mathbf{r}}

\makeatletter
\newcommand\footnoteref[1]{\protected@xdef\@thefnmark{\ref{#1}}\@footnotemark}
\makeatother
\usepackage{makecell} 

\def\ECCV18SubNumber{1288}
\title{ShapeStacks: Learning Vision-Based Physical Intuition for Generalised Object Stacking}

\titlerunning{ShapeStacks}
\authorrunning{Groth et al.}
\author{Oliver Groth, Fabian B. Fuchs, Ingmar Posner, Andrea Vedaldi}
\institute{Department of Engineering, University of Oxford}

\begin{document}
\pagestyle{headings}
\mainmatter
\maketitle

\begin{abstract}
Physical intuition is pivotal for intelligent agents to perform complex tasks. In this paper we investigate the passive acquisition of an intuitive understanding of physical principles as well as the active utilisation of this intuition in the context of generalised object stacking. To this end, we provide \emph{ShapeStacks}\footnote{Source code \& data are available at http://shapestacks.robots.ox.ac.uk}: a simulation-based dataset featuring 20,000 stack configurations composed of a variety of elementary geometric primitives richly annotated regarding semantics and structural stability. We train visual classifiers for binary stability prediction on the ShapeStacks data and scrutinise their learned physical intuition. Due to the richness of the training data our approach also generalises favourably to real-world scenarios achieving state-of-the-art stability prediction on a publicly available benchmark of block towers. We then leverage the physical intuition learned by our model to actively construct stable stacks and observe the emergence of an intuitive notion of \emph{stackability} - an inherent object affordance - induced by the active stacking task. Our approach performs well even in challenging conditions where it considerably exceeds the stack height observed during training or in cases where initially unstable structures must be stabilised via counterbalancing.
\keywords{Intuitive Physics, Stability Prediction, Object Stacking}
\end{abstract}

\section{Introduction}\label{s:intro}

\begin{figure}
	\centering
	\includegraphics[width=0.8\textwidth]{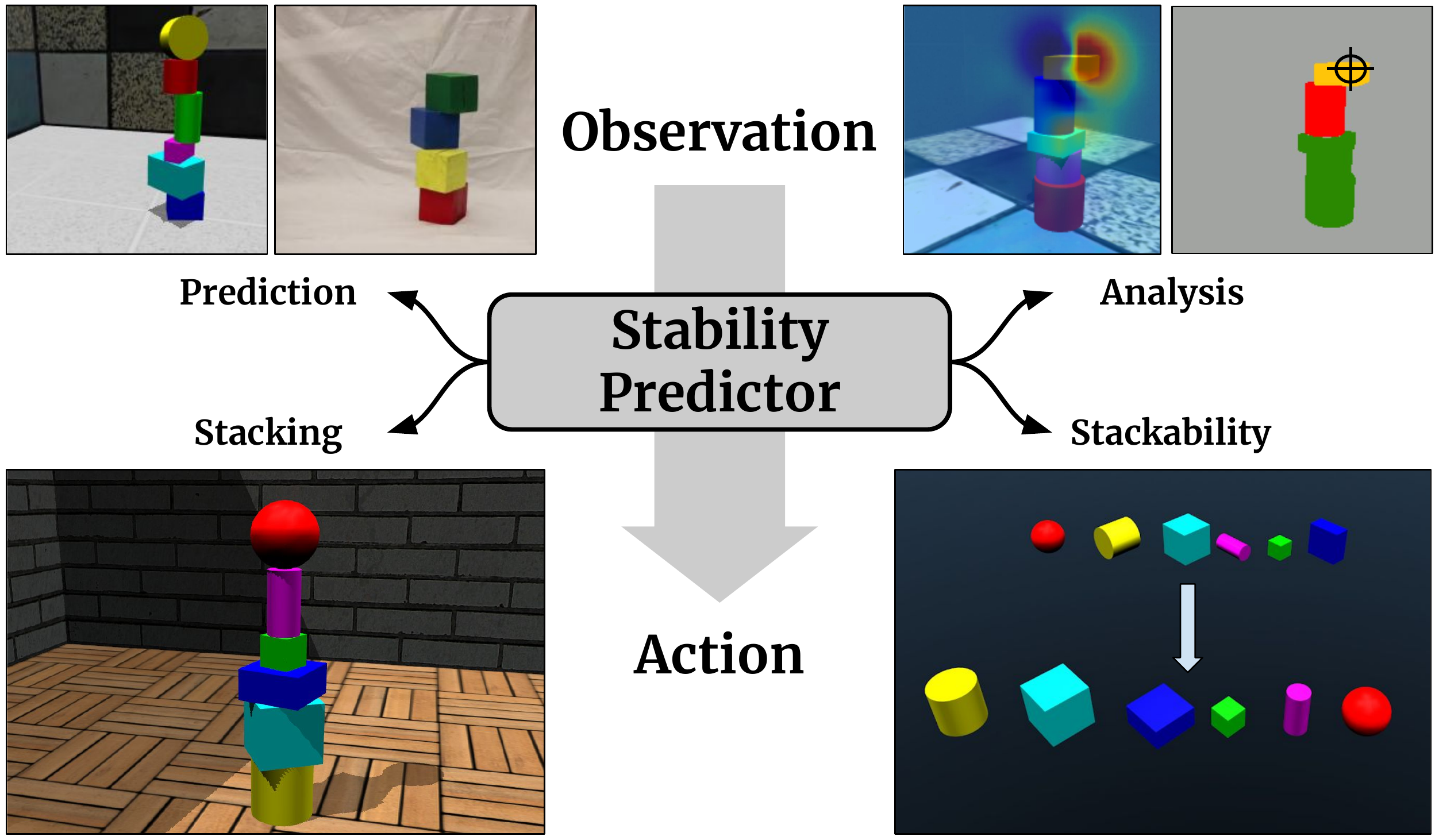}
	\caption{We present a visual classifier which is trained on stacks of diverse shapes to distinguish between stable and unstable structures. We demonstrate that the implicit knowledge captured by the predictor can be utilised to detect structural instabilities, infer the \emph{stackability} (utility with regard to stacking) of objects and guide a simulated stacking process solely from visual cues.}
	\label{fig:teaser}
\end{figure}

Research in cognitive science~\cite{Kubricht2017,Hamrick2016} highlights how the ability of humans to manipulate the environment depends strongly on our ability to intuitively understand its physics from visual observations. Intuitive physics may be just as important for autonomous agents to effectively and efficiently perform complex tasks such as object stacking or \mbox{(dis-)assembly} - and even the creation and use of tools. Central to these deliberations is an understanding of the physical properties of objects in the context of how they are meant to be used. Such object \emph{affordances} are typically pre-defined given knowledge of the task at hand \cite{Kjellstrom2011,Koppula2016}. In contrast, we posit that relevant affordances do not need to be specified a priori but can be learned in a task-driven manner.

Inspired by recent work in computer vision~\cite{Lerer2016,Wu2017} and robotics~\cite{Li2017,Furrer2017,YukeZhuZiyuWangJoshMerelAndreiRusuTomErezSerkanCabiSaranTunyasuvunakoolJanosKramarRaiaHadsellNandodeFreitas2018} we consider the task of \emph{object stacking} and the problem of learning -- from passive visual observations -- its intuitive physical principles.
By leveraging the model's acquired intuitions, we are able to utilise the passive observation in an active manipulation task as outlined in \Cref{fig:teaser}, which sets us apart from prior art in both scope and reach. 

Firstly, we argue that in order for agents to perform complex tasks they need to be able to interact with a variety of different object types. We therefore investigate the stacking problem using a broader set of geometric primitives than found in related works. To this end we introduce \emph{ShapeStacks}, a simulation-based dataset specifically created to enable exploration of stackability of a variety of objects. Furthermore, \emph{ShapeStacks} is, to the best of our knowledge, the first such dataset with annotations of the mechanical points of failure of stacks, which are inferred by formally analysing the underlying physics. This makes \emph{ShapeStacks} the most rigorous and complete publicly available dataset in this space.

Secondly, based on the \emph{ShapeStacks} dataset, we extend the investigation of stability prediction presented in~\cite{Lerer2016,Wu2017} to include stacks containing multiple object geometries. This allows for a more rigorous qualitative and quantitative evaluation of system performance. For example, our work, for the first time, quantifies if a model trained for stability prediction correctly localizes the underlying stability violations. We demonstrate that our model based on \emph{ShapeStacks} outperforms the baseline by Lerer et al.~\cite{Lerer2016} and performs commensurately with the current state-of-the-art \cite{Wu2017} on real-world image data without requiring a physics engine during test time.

Lastly, in order to investigate our main hypothesis -- namely that meaningful affordances emerge from representations learned by performing concrete tasks -- our work goes beyond the passive assessment of stacked towers as stable or unstable and actively performs stacking. In particular, we argue that, through the passive task of stability prediction, our system implicitly learns to assess the \emph{stackability} of the individual object geometries involved. We demonstrate this by extracting a stackability score for different block geometries and by using it to prioritise piece selection in the construction of tall stacks. By inserting noise in the actual stacking process in lieu of disturbances present in real agents (e.g. motor and perception noise as well as contact physics) we demonstrate that a more intuitive notion of object \emph{stackability} emerges.

As a result, our approach discovers an object's suitability towards stacking, ranks pieces accordingly and successfully builds stable towers. In addition, we show that our model is able to stabilise previously unstable structures by the addition of counterweights, arguably by developing an intuitive understanding of \emph{counterbalancing}.

\section{Related Work}\label{s:related}

The idea of vision-based physical intuition is firmly rooted in cognitive science where it is a long standing subject of investigation~\cite{Kubricht2017}. Humans are very apt at predicting structural stability~\cite{Battaglia2013}, inferring relative masses~\cite{Hamrick2016} and extrapolating trajectories of moving objects~\cite{Kubricht2017}. Although the exact workings of human physical intuition remain elusive, it has recently gained increasing traction in the machine learning, computer vision and robotics communities. The combination of powerful deep learning models and physics simulators yielded encouraging results in predicting the movement of objects on inclined surfaces~\cite{Wu2015} and the dynamics of ball collision~\cite{Fragkiadaki2015,Battaglia2016,Chang2016}.

While some prior work on intuitive physics assumed direct access to physical parameters, such as position and velocity, several authors have considered learning physics from visual observations instead. Examples include reasoning about support relations~\cite{GuptaEfrosHebert_ECCV10,Jia2015} and their geometric affordances and inferring forces in \emph{Newtonian} image understanding~\cite{Mottaghi2016}. Our aim is similar in that we learn the affordance of \emph{stackability} -- an object's utility towards stacking -- from visual observation. Importantly, however, in our work affordances are not specified \emph{a priori}, but emerge by passively predicting the stability of object stacks.

The latter is related to several recent works in stability prediction. Lerer et al. \cite{Lerer2016} pioneered the area by demonstrating feed-forward stability prediction of stacks from simulated and real images, releasing a collection of the latter as a public benchmark. Wu et al. \cite{Wu2017} proposed more sophisticated predictors based on re-rendering an observed scene and using a physics engine to compute stability, outperforming~\cite{Lerer2016} on their real-world data.
In contrast, our approach achieves performance commensurate to~\cite{Wu2017} while using only efficient feed forward prediction as in \cite{Lerer2016}.

The problem of structural stability is also well studied in the robotics community, especially in the context of manipulation tasks. Early work implements rule-based approaches with rudimentary visual perception for the game of Jenga \cite{Wang2009} or the safe deconstruction of object piles \cite{Ornan2013}.
More recently, advances in 3D perception and physical simulation have been exploited to stack irregular objects like stones \cite{Furrer2017}. 

The experimental setup of Li et al. \cite{Li2017,Li2016} is related to ours in that a stability predictor is trained for Kappla blocks in simulation which is then applied to guide stacking with a robotic arm. Our work is set apart from \cite{Li2017,Li2016} in that we are considering a variety of object geometries as well as more challenging stack configurations. Furthermore, \cite{Li2017,Li2016} do not consider object affordances.

More recently, Zhu et al. \cite{YukeZhuZiyuWangJoshMerelAndreiRusuTomErezSerkanCabiSaranTunyasuvunakoolJanosKramarRaiaHadsellNandodeFreitas2018} show that an end-to-end approach with an end-effector in the loop can be used to learn visuo-motor skills sufficient to stack two blocks on top of one another -- both in simulation and in the real world. Their work can be seen as complementary to ours, focusing on the end-effector actuation during stacking while we concentrate on the visual feedback loop and the emerging object affordances.

\section{The ShapeStacks Dataset}\label{s:method}

\begin{figure}[t]
	\centering
	\includegraphics[width=0.99\textwidth]{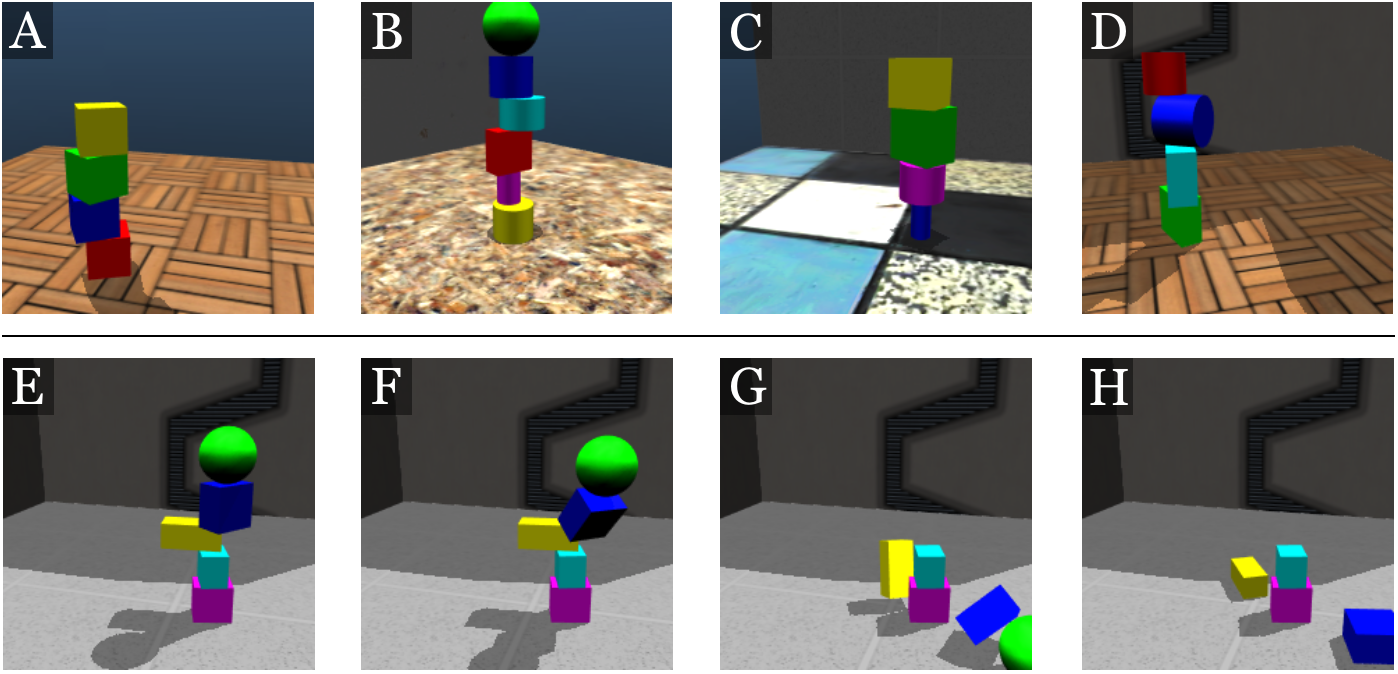}
	\caption{Different scenarios from the ShapeStacks data set. (A) - (D) depict initial stack setups: (A) stable, rectified tower of cubes, (B) stable tower where multiple objects counterbalance each other; some recorded images are cropped purposefully to include the difficulty of partial observability, (C) stable, but visually challenging scenario due to colours and textures, (D) violation of planar-surface-principle (VPSF). (E)~-~(H) show the simulation of an unstable, collapsing tower due to a centre of mass violation (VCOM).}
	\label{fig:scenario_types}
\end{figure}
\begin{table}[t]
\fontsize{8}{7.2}\selectfont 
\center
\caption{\emph{ShapeStacks} contents. On the left, we present the number of scenarios and recorded images in both subsets of the dataset. \emph{CCS} consists of cuboids, cylinders and spheres of varying size while \emph{Cubes} only features regular blocks. On the right, we report the rendering and annotation details. See \Cref{s:mechanics} for the derivation of the stability violation types \emph{VCOM} and \emph{VPSF}.}
\vspace{0.5cm}
\begin{subtable}{.68\textwidth}
\label{tab:dataset}
\center
\ra{1.3}
\begin{tabular}{@{}lrrrcrrrrr@{}}\toprule
\renewcommand{\arraystretch}{1.5}
& \multicolumn{3}{c}{\includegraphics[width=0.2\textwidth]{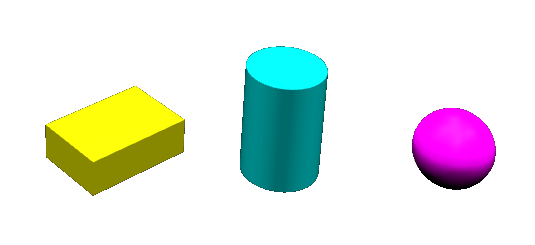}}
  && \multicolumn{3}{c}{\includegraphics[width=0.2\textwidth]{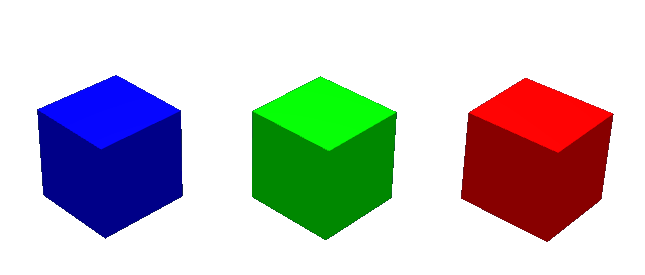}} \\ 
 & \multicolumn{3}{c}{\textbf{CCS} (\# Scenarios) }& \phantom{abc}& \multicolumn{3}{c}{\textbf{Cubes} (\# Scenarios)}   	& 
\\
\cmidrule{2-4} \cmidrule{6-8}
Stack height   & Train & Val & Test  && Train & Val & Test & \\ \midrule
$h=2$  & 1,340  & 286  & 286   && 1,680  & 360  & 360  & \\
$h=3$  & 2,464  & 528  & 528   && 1,680  & 360  & 360  & \\
$h=4$  & 1,716  & 368  & 368   && 1,558  & 332  & 332  & \\
$h=5$  & 678   & 144  & 144   && 1,274  & 272  & 272  & \\
$h=6$  & 194   & 40   & 40    && 1,030  & 220  & 220  & \\ \midrule
\# Scenarios  & 6,392  & 1,366 & 1,366  && 7,222  & 1,544 & 1,544 & \\
\# Images & 102,272 & 21,856 & 21,856 && 115,552 & 24,704 & 24,704 & \\

\bottomrule
\end{tabular}
\end{subtable}
\begin{subtable}{.02\textwidth}
\quad
\end{subtable}
\begin{subtable}{.28\textwidth}
\centering
\ra{1.5}
\setlength{\tabcolsep}{0.8em} 
\begin{tabular}{@{}l@{}}\toprule
\renewcommand{\arraystretch}{2.5}
Rendering \& Annotation \\ \midrule
\textbf{Rendering} \\
\checkmark $224 \times 224$ RGB \\
\textbf{Randomised Scenes} \\
\checkmark 25 Background Textures \\
\checkmark 6 Object Colours \\
\checkmark 5 Lighting Conditions \\
\textbf{Annotation} \\
\checkmark 0/1 Stability \\
\checkmark VCOM \& VPSF \\
\checkmark Scene Semantics \\ 
\bottomrule
\end{tabular}
\end{subtable}
\end{table}


In this section we describe the \emph{ShapeStacks} dataset, starting from an overview of its contents (\Cref{s:content}) followed by an analysis of the physics of stacking (\Cref{s:mechanics}).
The latter is required to explain the design of \emph{ShapeStacks} as well as to precisely define some of its physical data annotations.
The full dataset including simulation descriptions and data generation scripts is publicly available.

\subsection{Dataset Content}\label{s:content}

\emph{ShapeStacks} is a large collection of 20,000 simulated block-stacking scenarios. The selection of the scenarios emphasizes diversity by featuring multiple geometries, degrees of structural complexity and types of structural stability violations, as shown in~\Cref{fig:scenario_types}.

A detailed summary of the dataset content is provided in~\Cref{tab:dataset}. Each scenario is a single-stranded stack of cubes, cuboids, cylinders and spheres, all with varying dimensions, proportions and colours. The 20,000 scenarios are split roughly evenly among scenarios that contain only cubes (for comparing to related work on stability prediction~\cite{Lerer2016,Wu2017}, and scenarios containing cuboids, cylinders and spheres (abbrev.\ CCS). Stacks have variable heights, from two to six objects, with the majority built up to a height of three. Each scenario can either be stable or unstable. This is determined by running a physics simulation with the given scenario as starting condition\footnote{We only report and release scenarios where the simulation outcome aligns with the physical derivation. Scenarios which behave differently due to imprecisions of the simulator are discarded.}.
For every stack height, we provide an equal amount of stable and unstable scenarios.
Furthermore, unstable scenarios are evenly divided into the two different instability types (cf. ~\Cref{s:mechanics}).

Scenarios are split into train ($\sim 70\%$), validation ($\sim 15\%$), and test ($\sim 15\%$) sets.
Each scenario is rendered with a randomised set of background textures, object colours and lighting conditions.
We record every scenario from 16 different camera angles and save RGB images of a resolution of 224 x 224 pixels.

Every recorded image carries a binary stability label.
Also, every image is aligned with a segmentation map relating the 
different parts of the image to their semantics with regard to stability.
The segmentation map annotates the object which violates the stability of the tower, the first object to fall during the collapse and the base and top of the tower.


\subsection{The Mechanics of Stacking}\label{s:mechanics}
\label{sec:mechanics}
While our goal is to study intuitive physics and the emergence of object affordances, we argue that a precise understanding of the physical properties of the scenarios is essential to control data generation as well as to evaluate models.

In this paper, we restrict our attention to \emph{single-stranded stacks}: each object $S$ rests on top of another object $S'$ or the ground plane and no two objects are at the same level. That is, we exclude structures such as arches, multiple columns, forks, etc. We also assume that all objects are \emph{convex}, so that a straight line between any two points of the object is fully contained within it.

In order to determine the stability of a stack, we must use the notion of \emph{Centre of Mass} (CoM). Let $\bp = (x,y,z) \in S_i \subset \mathbb{R}^3$ be a point contained within the rigid body $S_i$. If $m$ is the mass of the object and if the material is homogeneous with density $\rho$, then its CoM is given by $\br_i = \rho \int_{S_i} \bp \,dx\,dy\,dz / m$.

We now study the stability of an object on top of another and then generalize the result to a full stack. For that, it is useful to refer to the topmost two blocks in~\Cref{f:com}. Assume that the rigid body $S_4$ is immersed in a uniform gravity field acting in the negative direction of the $z$ axis. Furthermore, assume that $S_4$ is resting on a horizontal surface (in this case $S_3$) such that all of its contact points are contained in a horizontal plane $\pi$ and $A \subset \pi$ denotes the convex hull of such points. Then $S_4$ is stable if, and only if, the projection of its CoM $\br_4$ on $\pi$ is contained in $A$~\cite{Wieber2002}, which we write as $\operatorname{Proj}_\pi(\br_4) \in A$. If $S_4$ rests in a stable position on $S_3$, the combination of ($S_3$, $S_4$) can be seen as a rigid body with CoM $\br_3^4$. We can then check the stability of the entity ($S_4$, $S_3$) with respect to $S_2$. Proceeding iteratively for every object from top to bottom of the stack results in the following lemma illustrated in~\Cref{f:com}:

\begin{figure}[t]
\centering\includegraphics[width=\textwidth]{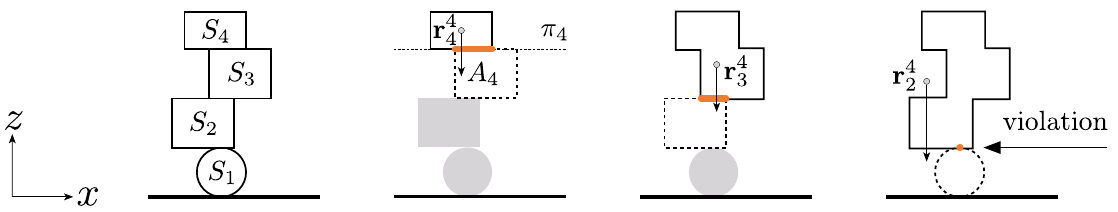}  
\caption{\textbf{Centre of Mass criterion.} The stability of a stack can be tested by considering sub-stacks sequentially, from top to bottom. For stability, the projection of the CoM of each sub-stack must lie within the contact surface with the block supporting it. As shown on the right, a cylindrical or spherical object offers an infinitesimally small contact surface which does not afford stability.}\label{f:com}
\end{figure}

\begin{lemma}\label{l:com}
Let $S_1,\dots,S_n$ be a collection of convex rigid bodies forming a single-stranded tower resting on a flat ground plane $S_0$. Let $m_1,\dots,m_n$ be the masses of the objects and $\br_1,\dots,\br_n$ their centres of mass.  Furthermore, let $A_{i}$ be the contact surface between object $S_{i-1}$ and $S_i$ and let $\pi_i \subset A_i$ be the plane containing it. Assume that $\pi$ is parallel to the $xy$ plane, which in turn is orthogonal to gravity. Then, if the objects are initially at rest, the tower is stable if, and only if,
\begin{equation}\label{eq:stability_criterion}
  \forall i=1,\dots,n-1:
  \quad
   \operatorname{Proj}_{\pi_i}(\br_{i+1}^n)
   \in A_i,
   \quad    
   \br_{i+1}^{n}
   =
   \frac{\sum_{j=i+1}^n m_j \br_j}
   {\sum_{j=i+1}^n m_j}
\end{equation}
where $\br_{i+1}^{n}$ is the overall CoM of the topmost $n-i$ blocks.
\end{lemma}

This lemma can be used to assess the stability of a stack by checking the CoM condition from top to bottom for every interface $A_i$. 
Note that what is important is not the centre of mass of the individual blocks, but that of the part of the tower above each surface $A_i$. 
Thus it is possible to construct a stable stack that has apparent CoM violations for individual blocks, but that is overall stable due to the counter-balancing effect of the other blocks on top.
Importantly, this allows for complex stacks that cannot be constructed in a bottom-up manner by placing only one object at a time.

We specifically distinguish between two types of instabilities.
The first is \emph{violation of the planar surface criterion} (VPSF). This is caused by an object stacked on top of a curved surface which violates~\Cref{eq:stability_criterion} due to the infinitesimally small contact area. It is worth noting that this depends on the shape of the objects and not on the relative object positioning.
The second type of instability is called \emph{violation of the centre of mass criterion} (VCOM), and comprises violations of~\Cref{eq:stability_criterion} that depend instead on the positioning of the objects in the stack.
For each unstable scenario we introduce either a VPSF or a VCOM violation for exactly one contact area $A_i$.

For dataset construction, \Cref{l:com} thus allows us to tightly control which stability violation occurs in each simulated scenario and to mark in each image which object it is attributable to (cf.~\Cref{fig:results_attention}).

\section{Stability Prediction}\label{s:experiments}

In this section, we construct models that can predict the stability of a stack from RGB images alone. We learn these models from passive observations of stable and unstable stacks. 
Specifically, our vision-based stability classifier is trained to distinguish between stable and unstable towers (\Cref{sec:stability_prediction}) and validated by demonstrating state-of-the-art performance on both simulated and real data.
We also quantify how reliably the models can localise the mechanical stability violations present in the unstable stacks (\Cref{sec:attention}).

\subsection{Training the Stability Predictor}\label{sec:stability_prediction}

We train a visual classifier for the task of predicting whether a shape stack is stable or not using images\footnote{We only use still images of initial stack configurations and no images depicting collapses from later time points in the simulations.} from the \emph{ShapeStacks} dataset, annotated with binary stability labels.

To this end we investigate the use of two neural network architectures commonly used for image-based classification: AlexNet~\cite{Krizhevsky2012} and Inception~v4~\cite{Szegedy}. In both cases we optimise the network parameters $\theta$ given our dataset $D = \{(x^{(1)}, y^{(1)}), \dots, (x^{(m)}, y^{(m)})\}$ of images $x^{(i)}$ and stability labels $y^{(i)}$ by minimising the following logistic regression loss:

\vspace{-0.3cm}
\begin{equation}\label{eq:loss}
	L(\theta; D) = 
    -\sum_{i=1}^{m}{y^{(i)} \log \left(\frac{1}{1 + e^{-f(x^{(i)}; \theta)}}\right) + (1-y^{(i)}) \log \left(1 - \frac{1}{1 + e^{-f(x^{(i)}; \theta)}}\right)}
\end{equation}

The unscaled logit output of the CNNs is denoted by $f(x;\theta)$ and the label values are $y=0$ for stable and $y=1$ for unstable images.
Inception v4 and AlexNet are both trained using the RMSProp optimiser \cite{Hinton2014} with solver hyper-parameters as reported in~\cite{Szegedy} for 80 epochs.

We use the two different subsets of \emph{ShapeStacks} during training (cf. \Cref{tab:dataset}), each one containing an equal amount of stable and unstable images.
Both types of violations (VCOM and VPSF, cf. \Cref{sec:mechanics}) are evenly represented among unstable images.
We also reserve a set of 46,560 images featuring stacks of all shapes as final test set.
During training, we augment the training images by randomising colours, varying aspect-ratios, and applying random cropping, vertical flipping and minimal in-plane rotation.
We ensure that all data augmentations still yield physically plausible, upright towers.

\Cref{tab:stability_prediction} presents the performance of the classifiers on our simulated test data and on the real-world block tower data provided by \cite{Lerer2016}.
Our experiments suggest that AlexNet provides a useful baseline for CNN performance on this task. However, it is consistently outperformed by the Inception network.
We choose the Inception v4 architecture trained on ShapeStacks data as the reference model in all further experiments.

As expected, both models perform best on the real-world data when only trained on cubes as the real-world images also only show stacks of cubes. Best performance is reached for both models on the combined ShapeStacks test data (featuring all shapes) when training is also performed on multiple object types.
However, it is surprising how well the Inception network generalises from cubes to other structures suggesting that it learned an intuition about the CoM principle (\cref{s:mechanics}) which is also applicable to more complex shapes.

On real images, Inception v4, trained from scratch on our dataset, outperforms the baseline from Lerer et al. \cite{Lerer2016} and is on par with the more complex visual de-animation approach by Wu et al. \cite{Wu2017}, which translates the observed images into a physical state and checks stability with a physics engine.
We attribute this to the richness of the \emph{ShapeStacks} dataset as well as to our data augmentation scheme, which results in a visually and structurally diverse set of stacks and hence affords good generalisation.

\begin{table}[t]
\caption{Stability prediction accuracy given as the percentage of correctly classified images into stable or unstable. AlexNet and Inception v4 (INCPv4) are trained from scratch on simulated data consisting of stacks featuring either cubes or CCS. INCPv4-IMGN is pre-trained on ImageNet~\cite{Deng2009}. All algorithms are tested on both \emph{real} images from \cite{Lerer2016} and \emph{simulated} images from our \emph{ShapeStacks} test split featuring all shapes.}
\label{tab:stability_prediction}
\center


\ra{1.3}
\setlength{\tabcolsep}{0.3em} 
\begin{tabular}{@{}lcrrcrrcrrcrr@{}}\toprule
\renewcommand{\arraystretch}{1.5}
& \phantom{a}& \multicolumn{2}{c}{AlexNet}  & \phantom{a}& \multicolumn{2}{c}{INCPv4-IMGN}& \phantom{a}& \multicolumn{2}{c}{\textbf{INCPv4}}& \phantom{a} & Physnet & VDA \\
\cmidrule{3-4} \cmidrule{6-7} \cmidrule{9-10}
&& Cubes & CCS && Cubes & CCS && Cubes & CCS &&\cite{Lerer2016} &  \cite{Wu2017}  \\ \midrule
Simulated && 60.5\% & 58.8\% && 76.2\% & \textbf{84.9\%} && 77.7\% & \textbf{84.9}\% && N/A\tablefootnote{\label{note_nocmp}No comparison possible because neither training data nor model are publicly available.} & N/A\footnoteref{note_nocmp} \\
Real~\cite{Lerer2016}&& 65.5\% & 52.5\% && 73.2\% & 64.9\% && 74.7\% & 66.3\% && 66.7\% & \textbf{75}\%  \\ 
\midrule

\end{tabular}
\begin{tabular}{@{}m{1.5cm}m{1.15cm}m{1.15cm}m{1.6cm}cm{1.3cm}m{1.15cm}m{1.15cm}m{1.15cm}rcrrr@{}}
Simulated Examples
& {\includegraphics[width=0.1\textwidth]{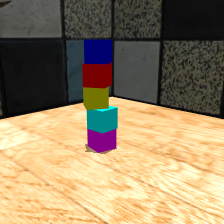}}
& {\includegraphics[width=0.1\textwidth]{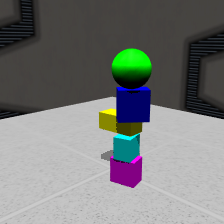}}
& {\includegraphics[width=0.1\textwidth]{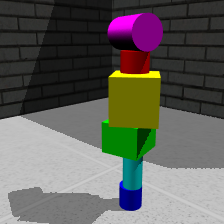}}
& \quad
& Real \mbox{Examples}
& {\includegraphics[width=0.1\textwidth]{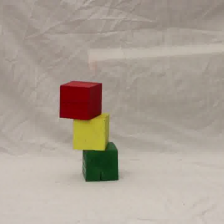}}
& {\includegraphics[width=0.1\textwidth]{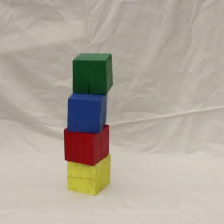}}
& {\includegraphics[width=0.1\textwidth]{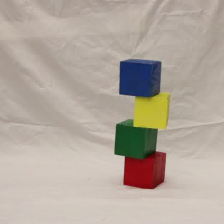}}
\\ \bottomrule
\end{tabular}
\end{table}


\subsection{Instability Localisation}
\label{sec:attention}
\begin{figure}[h!]
    \centering
    \includegraphics[width=1.0\textwidth]{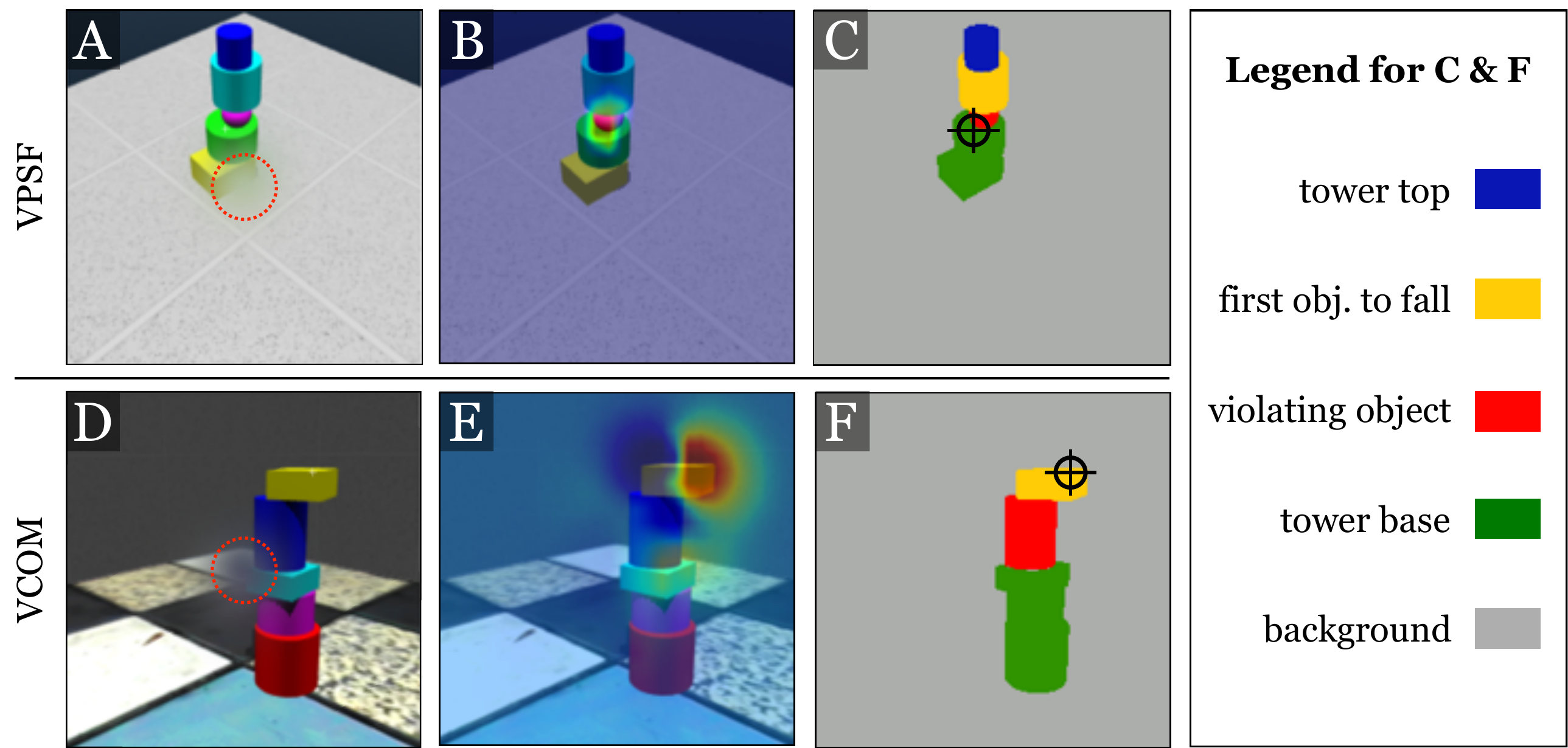}
	\caption{Attention visualisation obtained via an occlusion study.
    A Gaussian blur is applied in a sliding window manner to the image (A, D) , the increase (red) / decrease (blue) in the predicted stability is shown as a heatmap in (B, E), and the latter is compared to ground-truth segmentation maps in (C, F).
    The centres of attention are compared to the respective segmentation maps (C) and (F) and indeed correlate with the respective violation sites as indicated by the cross hairs.
}
	\label{fig:results_attention}
\end{figure}

\vspace{-0.3cm}
In order to probe whether the network grounds its stability prediction on sound mechanical principles we examine its ability to localise mechanical points of failure.
Our approach is similar to that of \cite{Lerer2016} though owing to the annotations included in the \emph{ShapeStacks} dataset we are able to conduct a quantitative analysis on 1,500 randomly sampled images from the test set by comparing the network's attention maps with the corresponding ground truth stability segmentation maps (cf. \Cref{fig:results_attention}).

Specifically, we compute the attention maps by conducting an occlusion study whereby images are blurred using a Gaussian filter with a standard deviation of 30 pixels applied in a sliding window manner with stride 8 and a patch size of 14 x 14 pixels. To avoid creating object-like occlusion artefacts, the blurred patch does not have rigid boundaries but gradually fades into the image (cf. \Cref{fig:results_attention}A and D).
The patched images are given as an input to the stability classifier and the predicted stability scores are aggregated in a map (cf. \Cref{fig:results_attention} B and E).

We then check whether the maximiser of the attention map is contained within the object responsible for stability violation (cf. \Cref{fig:results_attention} C and F) and report results in \Cref{tab:attention_distribution}.
In $79.9\%$ of all unstable cases, the network focuses on the violation region, which we define as the smallest rectangle enclosing the violating object and the first object to fall.

For VPSF instabilities, the network attends to the violating, curved object with a likelihood of $52.1\%$.
For VCOM instabilities, the network's main focus still remains on the violating object but is also spread out to the unsupported upper part of the tower (\emph{First Object to Fall} + \emph{Tower Top}) in $38.1\%$ of the cases, which is in line with the physics governing VCOM instabilities (cf. \cref{eq:stability_criterion}). 

\begin{table}[t]
\label{tab:attention_distribution}
\caption{The fraction of times the network attends image areas with specific physical meaning (cf.\ \Cref{fig:results_attention}). 1,500 images were analysed with an Inception v4 network trained on the CCS data (cf.\ \Cref{sec:stability_prediction}). The first row is aggregated over all instability types and the second and third rows offer a breakdown for the CoM (VCOM) and planar surface violations (VPSF), respectively. The fourth row lists the fractions of the areas occupied with the respective label across the segmentation maps of all unstable scenarios and serves as a reference point of how likely it is to focus on a specific area just by random chance. Likewise, the fifth row reports random chance attention within the tower.
}
\center
\ra{1.3}
\setlength{\tabcolsep}{.8em} 
\begin{tabular}{@{}rrrrrrr@{}}\toprule
\renewcommand{\arraystretch}{1.5}
 & Violating & First Obj. & Violation  & Tower   & Tower    & Back-      \\[-0.4em]
 & Object & to Fall & Area & Base & Top & ground \\
 \midrule
VCOM \&VPSF    & 38.9\%        & 29.3\% & 79.9\% & 5.9\%  & 5.5\%      & 20.4\% \\
VCOM           & 32.7\%        & 30.8\% & 76.5\% & 6.5\%  & 7.3\%      & 22.7\% \\
VPSF           & 52.1\%        & 26.3\% & 87.1\% & 4.6\%  & 1.7\%      & 15.4\% \\ \midrule
Random chance & 1.6\% & 1.9\% & 4.9\% & 1.7\% & 1.8\% & 93.0\% \\
Random in tower & 19.3\% & 22.9\% & 59.0\% & 20.5\% & 21.7\% & 14.5\% \\
\bottomrule
\end{tabular}
\end{table}

\section{Stacking and Stackability}\label{s:stackability}

So far, we have focussed on predicting the stability of stacks.
However, it is not \\
clear whether the models we learned understand the geometric affordances needed for actively building new stacks.

Here, we answer this question by considering three active stacking tasks. The first one is to estimate the \emph{stackability} of different objects and prioritise them while stacking (\Cref{sec:stackability}).
The second is to accurately estimate the optimal placement of blocks on a stack through visual feedback (\Cref{sec:stacking}).
The third is to counter-balance an unstable structure by placing an additional object on top (\Cref{sec:balancing}).
All tasks show encouraging performance indicating that models do indeed acquire actionable physical knowledge from passive stability prediction.

\vspace{-0.25cm}
\subsection{Stackability}\label{sec:stackability}

\begin{figure}[t]
    \centering    
	\begin{subfigure}[h]{\textwidth}
		\centering
		\includegraphics[width=0.99\textwidth]{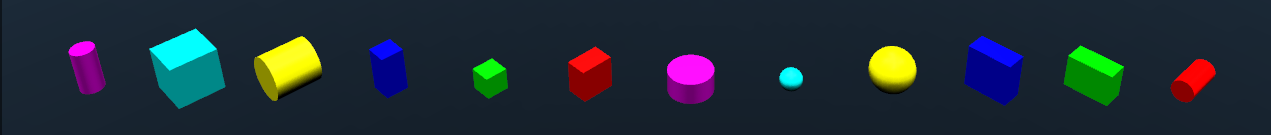}
		\label{fig:stackability_qualitative_1}
	\end{subfigure} 
	\begin{subfigure}[h]{\textwidth}
		\centering
		\includegraphics[width=0.99\textwidth]{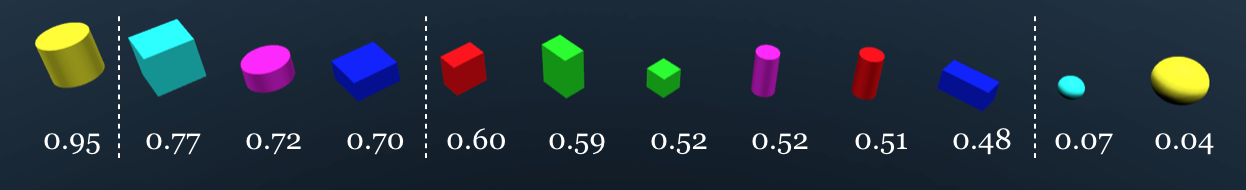}
		\label{fig:stackability_qualitative_3}
	\end{subfigure}
	\caption{\textit{Top row}: An unordered set of objects with random orientations. \textit{Bottom row}: objects sorted from most stackable (left) to least stackable (right). Every object is oriented in the way which affords best \emph{stackability} according to our network. The scores allow for division between different stability categories as visualised with white vertical lines.
    }
	\label{fig:stackability_qualitative}
\end{figure}

Different object shapes intrinsically have different stacking potential:
While a cuboid can serve as a solid base in every orientation, a cylinder can only support objects when placed upright and a sphere is never a good choice as a supporting object.
If an agent is given a set of blocks to stack, it can use an understanding of such affordances to prioritise objects, placing the most stable ones at the bottom of the stack.
We define \emph{stackability} of an object (i.e.\ its utility with regard to stack construction) by answering the question: ``How well can this object support the others in my set?''
Next, we show how to answer this question quantitatively using our learned stability predictor.

Given a set of objects, we compute their relative stackability scores as follows: Each object is placed on the ground as if it were the base of the stack using one of its discrete orientations\footnote{Cuboids afford three discrete orientations, one for each of its three distinct faces (considering symmetry). Cylinders afford two orientations (upright and sideways) and spheres afford only one orientation due to their radial symmetry.}.
Then, all other objects are systematically placed on top of the base object, one at a time, in all of their respective orientations. An image of the resulting combination is generated and assessed for stability using our predictor.
Positions for the top objects are sampled within a defined radius around the base object via simulated annealing and the maximum stability score is recorded.
The stackability score of the base object is then estimated as the average maximum stability achieved by all the other objects as they are placed on top of it. We also add random perturbations to the base position, with the idea of reflecting stackability robustness in the estimated score.

\begin{figure}[t]
    \centering
    \includegraphics[width=0.99\textwidth]{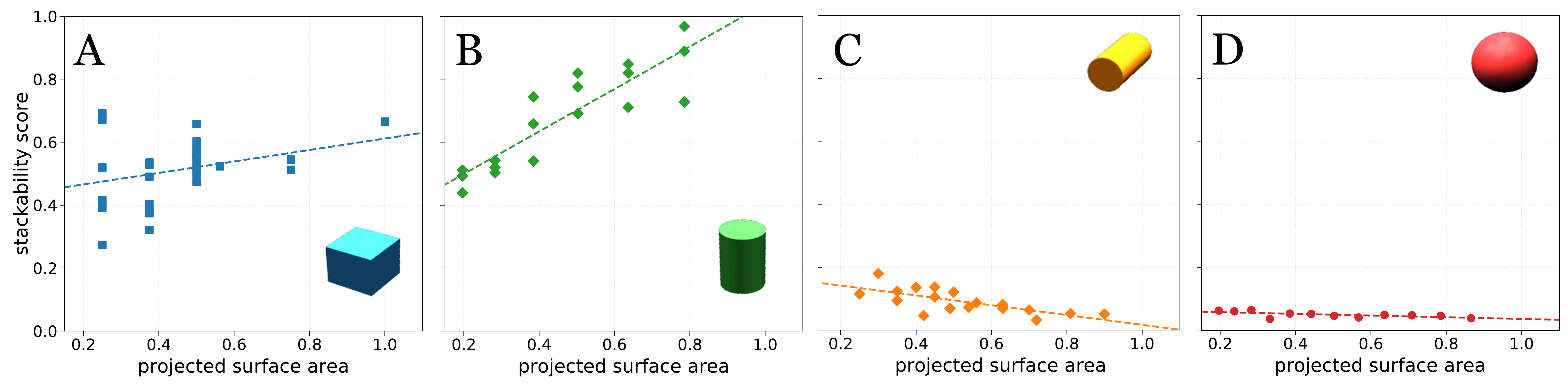}
	\caption{Correlation of the \emph{stackability} score with the projected surface area for different object classes. The projected surface area is calculated by projecting the object onto the x-y-plane. Spheres and lying cylinders are given very low \emph{stackability} scores. Upright cylinders and cuboids are generally more stackable as the projected surface area grows.}
	\label{fig:stackability_quantitative}
\end{figure}

Stackability can then be used to rank objects' shapes and orientations based on how well they can be expected to support other objects, as illustrated in~\Cref{fig:stackability_qualitative}.
We also examine the model's understanding of stackability quantitatively in \Cref{fig:stackability_quantitative} computing scores over all object classes with varying volumes and aspect ratios.
We generally find that the model ranks shapes in a sensible manner, preferring to stack on the largest face of cuboids, then on upright cylinders, and reject spheres as generally unsuitable for stacking.
The results suggest that the suitability of different geometries to stacking is implicitly learned by stability prediction.

\vspace{-0.3cm}
\subsection{Stacking Shapes in Simulation}\label{sec:stacking}

\begin{figure}[t]
    \centering
    \includegraphics[width=0.99\textwidth]{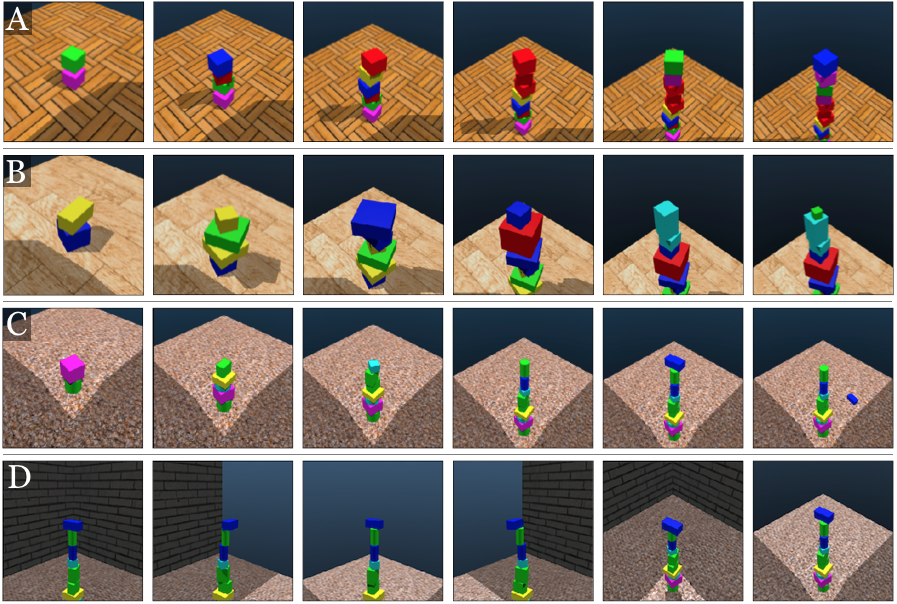}
	\caption{Three examples of stacking attempts. In (A) and (B), the algorithm successfully stacked up cubes and cuboids to the maximum height of 12. In C, the algorithm placed the 10th object in a way that violates \cref{eq:stability_criterion}. In (D), the images obtained from the different camera angles are shown for the failed stacking attempt in (C).}
	\label{fig:results_stacking_qual}
\end{figure}

Next, we investigate the ability of the stability predictor to not only order objects in an active stacking scenario, but also to accurately position them in stable configurations.
To do so, we design three stacking scenarios involving different shape types: cubes, cuboids and CCS. In each scenario, the method is given a pool of 12 different object shapes and sizes to stack with the goal of building as tall a tower as possible.
Every scenario is observed from six cameras (cf.~\Cref{fig:results_stacking_qual}D) which move upwards as the stack grows to guarantee full coverage of the process at any time.
At the beginning of every stacking episode, background textures, object colors and scene lights are randomised.
Then the stack order and best orientation for each objects are computed according to the stackability score (cf.~\Cref{sec:stackability}).

The stacking process commences with the first object being placed at the scene centre.
The object at place $r$ in the stacking queue is always spawned at a fixed height $h_r$ above the current tower trunk and candidate positions are sampled in the x-y-plane at $z=h_r$ according to the simulated annealing process described in \Cref{sec:stackability}.
If no stable position is identified for a particular object (i.e. logistic regression score $<0.5$), it is put aside and disregarded for the rest of the process.
The process is iterated until the placement of an object results in the collapse of the stack or no more objects are available.

\begin{figure}[t]
    \centering
    \includegraphics[width=0.99\textwidth]{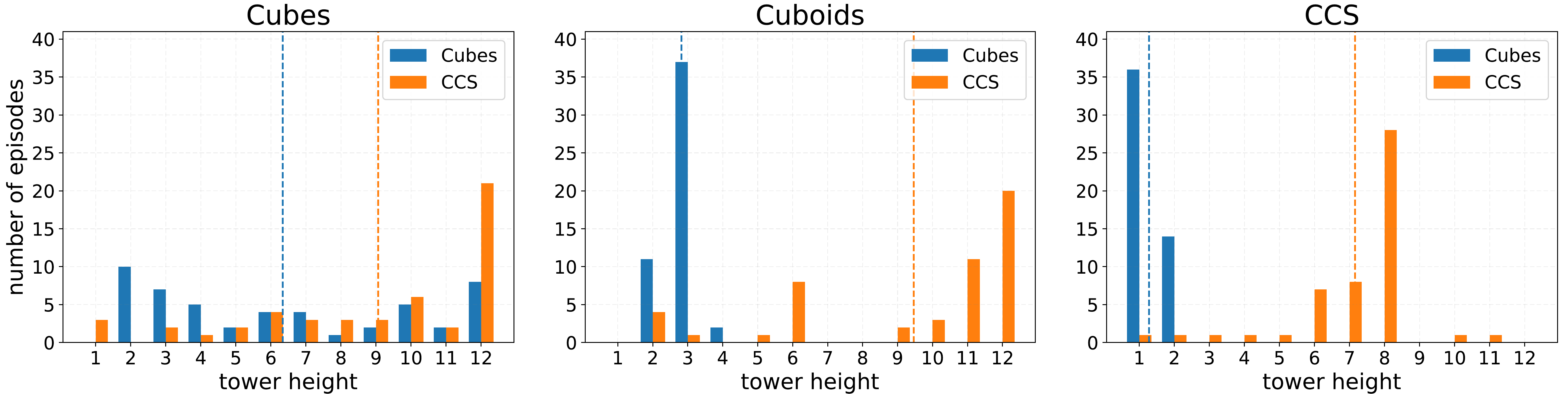}
	\caption{Stacking performance. The height of the bars indicate how often the algorithm built a tower with the respective number of  objects before it fell over. The mean tower height is indicated with a vertical dashed line.}
    \label{fig:results_stacking_quant}
\end{figure}

In \Cref{fig:results_stacking_quant}, we report achieved stack heights for two differently trained models in the three scenarios with cubes, cuboids and CCS, respectively. For each stacking episode, the algorithm is given a pool of 12 randomised objects. However, CCS scenarios always include exactly two spheres, so the maximum achievable height in this case is 11. We compare two stability predictors: One trained on cubes only (blue bars) and one trained on CSS objects (orange bars).
The CCS stability predictor clearly outperforms the one trained on cubes only in all three scenarios.
In fact, the cubes predictor only manages perform decently on cube stacking and largely fails when confronted with varied shapes highlighting the importance of training on a diverse shape set.

\vspace{-0.25cm}
\subsection{Balancing Unstable Structures}\label{sec:balancing}

\begin{figure}[t]
	\vspace{-0.3cm}
    \centering
    \begin{subfigure}[h]{0.68\textwidth}
    	\vspace{0.4cm}
		\centering
  		\includegraphics[width=.99\textwidth]{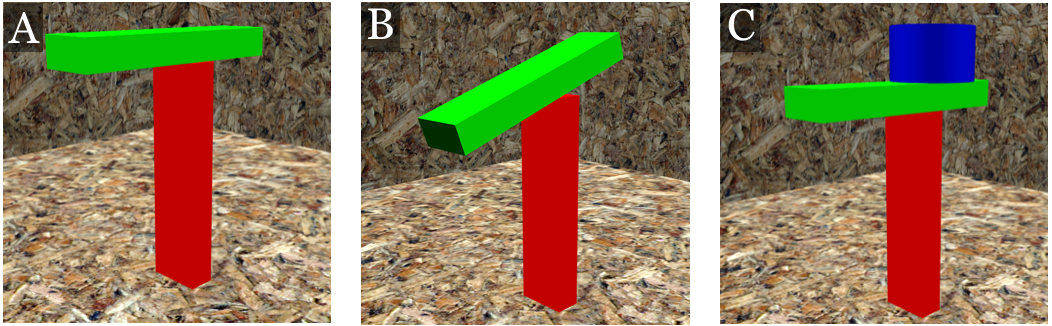}
		\label{fig:stackability_qualitative_2}
	\end{subfigure}
	\begin{subfigure}[h]{0.3\textwidth}
        \center
        \ra{1.1}
        \setlength{\tabcolsep}{.8em} 
        \begin{tabular}{@{}lr@{}}\toprule
         Object & Success Rate  \\
         \midrule
        Cube    & 76\% \\
        Cuboid           & 94\%\\
        Cylinder         & 72\%\\
        Sphere         & 98\%\\
        \bottomrule
        \end{tabular}
	\end{subfigure}
    \vspace{-0.3cm}
    \caption{Counterbalancing unstable structures. A: frozen, unstable stack; B: collapsing tower; C: successful placement of a counterweight that prevents collapse. Right: success rates for different counterweight types aggregated over 50 episodes.}
    \vspace{-0.2cm}
	\label{fig:balancing}
\end{figure}

In the final task, we present our model with an unstable stack, freeze it such that it does not collapse, and then ask the algorithm to place an additional object on top to counter-balance the instability. This is a subtle task that requires the model to understand the concept of counterbalancing and cannot be solved by simply centering a block on top of the one below.
\Cref{fig:balancing} shows that our algorithm successfully solves this task with high probability in an ``unstable T scenario'' for different types of counterweight objects.

\section{Conclusions}\label{s:conc}

We investigate the acquisition of physical intuition and geometric affordances in the context of vision-based, generalised object stacking.
To that end, we construct the \emph{ShapeStacks} dataset featuring diverse stacks of shapes with detailed annotations of mechanical stability violations and release it publicly.
We train a visual stability predictor on ShapeStacks which performs commensurately with state-of-the-art on simulated and real world images.
Our model also correctly localises structural instabilities, yields an intuitive notion about the stackability of objects and successfully guides a simulated stacking process solely based on visual cues.
Our results suggest that an intuitive understanding about physical principles and geometric affordances can be acquired from visual observation and effectively utilised in manipulation tasks.

\section*{Acknowledgements}
This research was funded by the European Research Council under grant ERC 677195-IDIU and the EPSRC AIMS Centre for Doctoral Training at Oxford University. We would like to thank Markus Wulfmeier for helpful comments on the draft of this paper. We also thank Adrian Penate-Sanchez, Rob Weston and Georgi Tinchev for proofreading.

\clearpage
\bibliographystyle{splncs}
\bibliography{refs}

\begin{thebibliography}{10}

\bibitem{Kubricht2017}
Kubricht, J.R., Holyoak, K.J., Lu, H.:
\newblock {Intuitive Physics: Current Research and Controversies}.
\newblock Trends in Cognitive Sciences \textbf{21}(10) (2017)  749--759

\bibitem{Hamrick2016}
Hamrick, J.B., Battaglia, P.W., Griffiths, T.L., Tenenbaum, J.B.:
\newblock {Inferring mass in complex scenes by mental simulation}.
\newblock Cognition \textbf{157} (2016)

\bibitem{Kjellstrom2011}
Kjellstr{\"{o}}m, H., Romero, J., Kragi{\'{c}}, D.:
\newblock {Visual object-action recognition: Inferring object affordances from
  human demonstration}.
\newblock Computer Vision and Image Understanding \textbf{115}(1) (2011)
  81--90

\bibitem{Koppula2016}
Koppula, H.S., Saxena, A.:
\newblock {Anticipating Human Activities Using Object Affordances for Reactive
  Robotic Response}.
\newblock IEEE Transactions on Pattern Analysis and Machine Intelligence
  \textbf{38}(1) (2016)  14--29

\bibitem{Lerer2016}
Lerer, A., Gross, S., Fergus, R.:
\newblock Learning physical intuition of block towers by example.
\newblock In: Proceedings of the 33rd International Conference on International
  Conference on Machine Learning - Volume 48. ICML'16, JMLR.org (2016)
  430--438

\bibitem{Wu2017}
Wu, J., Lu, E., Kohli, P., Freeman, W.T., Tenenbaum, J.B.:
\newblock {Learning to See Physics via Visual De-animation}.
\newblock Advances in Neural Information Processing Systems (Nips) (2017)

\bibitem{Li2017}
Li, W., Leonardis, A., Fritz, M.:
\newblock {Visual stability prediction for robotic manipulation}.
\newblock Proceedings - IEEE International Conference on Robotics and
  Automation (2017)  2606--2613

\bibitem{Furrer2017}
Furrer, F., Wermelinger, M., Yoshida, H., Gramazio, F., Kohler, M., Siegwart,
  R., Hutter, M.:
\newblock {Autonomous robotic stone stacking with online next best object
  target pose planning}.
\newblock Proceedings - IEEE International Conference on Robotics and
  Automation (2017)  2350--2356

\bibitem{YukeZhuZiyuWangJoshMerelAndreiRusuTomErezSerkanCabiSaranTunyasuvunakoolJanosKramarRaiaHadsellNandodeFreitas2018}
Zhu, Y., Wang, Z., Merel, J., Rusu, A.A., Erez, T., Cabi, S., Tunyasuvunakool,
  S., Kram{\'{a}}r, J., Hadsell, R., de~Freitas, N., Heess, N.:
\newblock Reinforcement and imitation learning for diverse visuomotor skills.
\newblock CoRR \textbf{abs/1802.09564} (2018)

\bibitem{Battaglia2013}
Battaglia, P.W., Hamrick, J.B., Tenenbaum, J.B.:
\newblock {Simulation as an engine of physical scene understanding}.
\newblock Proceedings of the National Academy of Sciences \textbf{110}(45)
  (2013)  18327--18332

\bibitem{Wu2015}
Wu, J., Yildirim, I., Lim, J., Freeman, W., Tenenbaum, J.:
\newblock {Galileo : Perceiving Physical Object Properties by Integrating a
  Physics Engine with Deep Learning}.
\newblock Advances in Neural Information Processing Systems 28 (NIPS 2015)
  (2015)  1--9

\bibitem{Fragkiadaki2015}
Fragkiadaki, K., Agrawal, P., Levine, S., Malik, J.:
\newblock {Learning Visual Predictive Models of Physics for Playing Billiards}.
\newblock (2015)  1--12

\bibitem{Battaglia2016}
Battaglia, P., Pascanu, R., Lai, M., Rezende, D.J.,  et~al.:
\newblock Interaction networks for learning about objects, relations and
  physics.
\newblock In: Advances in neural information processing systems. (2016)
  4502--4510

\bibitem{Chang2016}
Chang, M.B., Ullman, T., Torralba, A., Tenenbaum, J.B.:
\newblock {A Compositional Object-Based Approach to Learning Physical
  Dynamics}.
\newblock (2016)  1--15

\bibitem{GuptaEfrosHebert_ECCV10}
Gupta, A., Efros, A.A., Hebert, M.:
\newblock Blocks world revisited: Image understanding using qualitative
  geometry and mechanics.
\newblock In: European Conference on Computer Vision(ECCV). (2010)

\bibitem{Jia2015}
Jia, Z., Gallagher, A.C., Saxena, A., Chen, T.:
\newblock {3D reasoning from blocks to stability}.
\newblock IEEE Transactions on Pattern Analysis and Machine Intelligence
  \textbf{37}(5) (2015)  905--918

\bibitem{Mottaghi2016}
Mottaghi, R., Bagherinezhad, H., Rastegari, M., Farhadi, A.:
\newblock {Newtonian Image Understanding: Unfolding the Dynamics of Objects in
  Static Images}.
\newblock In: 2016 IEEE Conference on Computer Vision and Pattern Recognition
  (CVPR). (2016)

\bibitem{Wang2009}
Wang, J., Rogers, P., Parker, L., Brooks, D., Stilman, M.:
\newblock {Robot Jenga: Autonomous and strategic block extraction}.
\newblock 2009 IEEE/RSJ International Conference on Intelligent Robots and
  Systems, IROS 2009 (2009)  5248--5253

\bibitem{Ornan2013}
Ornan, O., Degani, A.:
\newblock {Toward autonomous disassembling of randomly piled objects with
  minimal perturbation}.
\newblock IEEE International Conference on Intelligent Robots and Systems
  (2013)  4983--4989

\bibitem{Li2016}
Li, W., Azimi, S., Leonardis, A., Fritz, M.:
\newblock To fall or not to fall: A visual approach to physical stability
  prediction.
\newblock arXiv preprint arXiv:1604.00066 (2016)

\bibitem{Wieber2002}
Wieber, P.B.:
\newblock {On the stability of walking systems}.
\newblock Proceedings of the Third IARP International Workshop on Humanoid and
  Human Friendly Robotics (2002)  1--7

\bibitem{Krizhevsky2012}
Krizhevsky, A., Sutskever, I., Hinton, G.E.:
\newblock {ImageNet Classification with Deep Convolutional Neural Networks}.
\newblock Advances In Neural Information Processing Systems (2012)  1--9

\bibitem{Szegedy}
Szegedy, C., Ioffe, S., Vanhoucke, V., Alemi, A.A.:
\newblock Inception-v4, inception-resnet and the impact of residual connections
  on learning.
\newblock In: AAAI. Volume~4. (2017) ~12

\bibitem{Hinton2014}
Hinton, G., Srivastava, N., Swersky, K.:
\newblock Coursera, neural networks for machine learning, lecture 6e.
\newblock (2014)

\bibitem{Deng2009}
Deng, J., Dong, W., Socher, R., Li, L.J., Li, K., Fei-Fei, L.:
\newblock Imagenet: A large-scale hierarchical image database.
\newblock In: Computer Vision and Pattern Recognition, 2009. CVPR 2009. IEEE
  Conference on, IEEE (2009)  248--255

\end{thebibliography}

\end{document}